\documentclass[runningheads]{llncs}

\usepackage[numbers,sort&compress]{natbib}
\usepackage[T1]{fontenc}    
\usepackage{microtype}
\usepackage{graphicx}
\usepackage[small]{caption}
\usepackage{subcaption}
\usepackage{booktabs} 
\usepackage[hidelinks]{hyperref}

\usepackage{amsmath}
\usepackage{amsfonts}       
\usepackage{amssymb}
\usepackage{mathtools}
\usepackage{nicefrac}       

\usepackage{paralist}

\usepackage[capitalize,noabbrev]{cleveref}
\usepackage[textsize=tiny]{todonotes}

\usepackage{times}

\usepackage{soul}
\usepackage{url}
\usepackage[utf8]{inputenc}

\usepackage{outlines}
\usepackage{optidef}
\usepackage[per-mode=symbol]{siunitx}
\usepackage{xcolor}

\newcommand{\vect}[1]{\mathbf{#1}} 
\newcommand{\vects}[1]{\boldsymbol{#1}} 

\usepackage{pgfplots}
\usepgfplotslibrary{groupplots}

\pgfplotsset{compat=newest}
\pgfplotsset{every axis legend/.append style={legend cell align=left}}

\pgfplotsset{every axis/.append style={
                    title style={font=\small},
                    tick label style={font=\footnotesize}  
                    }}
\pgfplotsset{every axis label/.style={font=\small}}                    
\pgfplotsset{
legend image code/.code={
\draw[mark repeat=2,mark phase=2]
plot coordinates {
(0cm,0cm)
(0.15cm,0cm)        
(0.3cm,0cm)         
};%
}
}

\definecolor{pastelMagenta}{HTML}{FF48CF}
\definecolor{pastelPurple}{HTML}{8770FE}
\definecolor{pastelBlue}{HTML}{1BA1EA}
\definecolor{pastelSeaGreen}{HTML}{14B57F}
\definecolor{pastelGreen}{HTML}{3EAA0D}
\definecolor{pastelOrange}{HTML}{C38D09}
\definecolor{pastelRed}{HTML}{F5615C}
\definecolor{lightGray}{RGB}{169, 169, 169}
\definecolor{darkRed}{RGB}{139, 0, 0}

\pgfplotsset{
   	colormap={pasteljet}{
		rgb=(0.99325,0.90616,0.14394)
		rgb=(0.98387,0.90487,0.13690)
		rgb=(0.97442,0.90359,0.13021)
		rgb=(0.96489,0.90232,0.12394)
		rgb=(0.95530,0.90107,0.11813)
		rgb=(0.94564,0.89982,0.11284)
		rgb=(0.93590,0.89857,0.10813)
		rgb=(0.92611,0.89733,0.10407)
		rgb=(0.91624,0.89609,0.10072)
		rgb=(0.90631,0.89485,0.09813)
		rgb=(0.89632,0.89362,0.09634)
		rgb=(0.88627,0.89237,0.09537)
		rgb=(0.87617,0.89112,0.09525)
		rgb=(0.86601,0.88987,0.09595)
		rgb=(0.85581,0.88860,0.09745)
		rgb=(0.84556,0.88732,0.09970)
		rgb=(0.83527,0.88603,0.10265)
		rgb=(0.82494,0.88472,0.10622)
		rgb=(0.81458,0.88339,0.11035)
		rgb=(0.80418,0.88205,0.11496)
		rgb=(0.79376,0.88068,0.12001)
		rgb=(0.78331,0.87928,0.12540)
		rgb=(0.77285,0.87787,0.13111)
		rgb=(0.76237,0.87642,0.13706)
		rgb=(0.75188,0.87495,0.14323)
		rgb=(0.74139,0.87345,0.14956)
		rgb=(0.73089,0.87192,0.15603)
		rgb=(0.72039,0.87035,0.16260)
		rgb=(0.70990,0.86875,0.16926)
		rgb=(0.69942,0.86712,0.17597)
		rgb=(0.68894,0.86545,0.18272)
		rgb=(0.67849,0.86374,0.18950)
		rgb=(0.66805,0.86200,0.19629)
		rgb=(0.65764,0.86022,0.20308)
		rgb=(0.64726,0.85840,0.20986)
		rgb=(0.63690,0.85654,0.21662)
		rgb=(0.62658,0.85464,0.22335)
		rgb=(0.61629,0.85271,0.23005)
		rgb=(0.60604,0.85073,0.23671)
		rgb=(0.59584,0.84872,0.24333)
		rgb=(0.58568,0.84666,0.24990)
		rgb=(0.57556,0.84457,0.25642)
		rgb=(0.56550,0.84243,0.26288)
		rgb=(0.55548,0.84025,0.26928)
		rgb=(0.54552,0.83804,0.27563)
		rgb=(0.53562,0.83579,0.28191)
		rgb=(0.52578,0.83349,0.28813)
		rgb=(0.51599,0.83116,0.29428)
		rgb=(0.50627,0.82879,0.30036)
		rgb=(0.49661,0.82638,0.30638)
		rgb=(0.48703,0.82393,0.31232)
		rgb=(0.47750,0.82144,0.31820)
		rgb=(0.46805,0.81892,0.32400)
		rgb=(0.45867,0.81636,0.32973)
		rgb=(0.44937,0.81377,0.33538)
		rgb=(0.44014,0.81114,0.34097)
		rgb=(0.43098,0.80847,0.34648)
		rgb=(0.42191,0.80577,0.35191)
		rgb=(0.41291,0.80304,0.35727)
		rgb=(0.40400,0.80027,0.36255)
		rgb=(0.39517,0.79748,0.36776)
		rgb=(0.38643,0.79464,0.37289)
		rgb=(0.37778,0.79178,0.37794)
		rgb=(0.36921,0.78889,0.38291)
		rgb=(0.36074,0.78596,0.38781)
		rgb=(0.35236,0.78301,0.39264)
		rgb=(0.34407,0.78003,0.39738)
		rgb=(0.33588,0.77702,0.40205)
		rgb=(0.32780,0.77398,0.40664)
		rgb=(0.31981,0.77091,0.41115)
		rgb=(0.31193,0.76782,0.41559)
		rgb=(0.30415,0.76470,0.41994)
		rgb=(0.29648,0.76156,0.42422)
		rgb=(0.28892,0.75839,0.42843)
		rgb=(0.28148,0.75520,0.43255)
		rgb=(0.27415,0.75199,0.43660)
		rgb=(0.26694,0.74875,0.44057)
		rgb=(0.25986,0.74549,0.44447)
		rgb=(0.25290,0.74221,0.44828)
		rgb=(0.24607,0.73891,0.45202)
		rgb=(0.23937,0.73559,0.45569)
		rgb=(0.23281,0.73225,0.45928)
		rgb=(0.22640,0.72889,0.46279)
		rgb=(0.22012,0.72551,0.46623)
		rgb=(0.21400,0.72211,0.46959)
		rgb=(0.20803,0.71870,0.47287)
		rgb=(0.20222,0.71527,0.47608)
		rgb=(0.19657,0.71183,0.47922)
		rgb=(0.19109,0.70837,0.48228)
		rgb=(0.18578,0.70489,0.48527)
		rgb=(0.18065,0.70140,0.48819)
		rgb=(0.17571,0.69790,0.49103)
		rgb=(0.17095,0.69438,0.49380)
		rgb=(0.16638,0.69086,0.49650)
		rgb=(0.16202,0.68732,0.49913)
		rgb=(0.15785,0.68376,0.50169)
		rgb=(0.15389,0.68020,0.50417)
		rgb=(0.15015,0.67663,0.50659)
		rgb=(0.14662,0.67305,0.50894)
		rgb=(0.14330,0.66946,0.51121)
		rgb=(0.14021,0.66586,0.51343)
		rgb=(0.13734,0.66225,0.51557)
		rgb=(0.13469,0.65864,0.51765)
		rgb=(0.13227,0.65501,0.51966)
		rgb=(0.13007,0.65138,0.52161)
		rgb=(0.12809,0.64775,0.52349)
		rgb=(0.12633,0.64411,0.52531)
		rgb=(0.12478,0.64046,0.52707)
		rgb=(0.12344,0.63681,0.52876)
		rgb=(0.12231,0.63315,0.53040)
		rgb=(0.12138,0.62949,0.53197)
		rgb=(0.12064,0.62583,0.53349)
		rgb=(0.12008,0.62216,0.53495)
		rgb=(0.11970,0.61849,0.53635)
		rgb=(0.11948,0.61482,0.53769)
		rgb=(0.11942,0.61114,0.53898)
		rgb=(0.11951,0.60746,0.54022)
		rgb=(0.11974,0.60379,0.54140)
		rgb=(0.12009,0.60010,0.54253)
		rgb=(0.12057,0.59642,0.54361)
		rgb=(0.12115,0.59274,0.54464)
		rgb=(0.12183,0.58905,0.54562)
		rgb=(0.12261,0.58537,0.54656)
		rgb=(0.12346,0.58169,0.54744)
		rgb=(0.12440,0.57800,0.54829)
		rgb=(0.12539,0.57432,0.54909)
		rgb=(0.12645,0.57063,0.54984)
		rgb=(0.12757,0.56695,0.55056)
		rgb=(0.12873,0.56327,0.55123)
		rgb=(0.12993,0.55958,0.55186)
		rgb=(0.13117,0.55590,0.55246)
		rgb=(0.13244,0.55222,0.55302)
		rgb=(0.13374,0.54853,0.55354)
		rgb=(0.13507,0.54485,0.55403)
		rgb=(0.13641,0.54117,0.55448)
		rgb=(0.13777,0.53749,0.55491)
		rgb=(0.13915,0.53381,0.55530)
		rgb=(0.14054,0.53013,0.55566)
		rgb=(0.14194,0.52645,0.55599)
		rgb=(0.14334,0.52277,0.55629)
		rgb=(0.14476,0.51909,0.55657)
		rgb=(0.14618,0.51541,0.55682)
		rgb=(0.14761,0.51173,0.55705)
		rgb=(0.14904,0.50805,0.55725)
		rgb=(0.15048,0.50437,0.55743)
		rgb=(0.15192,0.50069,0.55759)
		rgb=(0.15336,0.49700,0.55772)
		rgb=(0.15482,0.49331,0.55784)
		rgb=(0.15627,0.48962,0.55794)
		rgb=(0.15773,0.48593,0.55801)
		rgb=(0.15919,0.48224,0.55807)
		rgb=(0.16067,0.47854,0.55812)
		rgb=(0.16214,0.47484,0.55814)
		rgb=(0.16362,0.47113,0.55815)
		rgb=(0.16512,0.46742,0.55814)
		rgb=(0.16662,0.46371,0.55812)
		rgb=(0.16813,0.45999,0.55808)
		rgb=(0.16965,0.45626,0.55803)
		rgb=(0.17118,0.45253,0.55797)
		rgb=(0.17272,0.44879,0.55788)
		rgb=(0.17427,0.44504,0.55779)
		rgb=(0.17584,0.44129,0.55768)
		rgb=(0.17742,0.43753,0.55756)
		rgb=(0.17902,0.43376,0.55743)
		rgb=(0.18063,0.42997,0.55728)
		rgb=(0.18226,0.42618,0.55712)
		rgb=(0.18390,0.42238,0.55694)
		rgb=(0.18556,0.41857,0.55675)
		rgb=(0.18723,0.41475,0.55655)
		rgb=(0.18892,0.41091,0.55633)
		rgb=(0.19063,0.40706,0.55609)
		rgb=(0.19236,0.40320,0.55584)
		rgb=(0.19410,0.39932,0.55556)
		rgb=(0.19586,0.39543,0.55528)
		rgb=(0.19764,0.39153,0.55497)
		rgb=(0.19943,0.38761,0.55464)
		rgb=(0.20124,0.38367,0.55429)
		rgb=(0.20306,0.37972,0.55393)
		rgb=(0.20490,0.37575,0.55353)
		rgb=(0.20676,0.37176,0.55312)
		rgb=(0.20862,0.36775,0.55268)
		rgb=(0.21050,0.36373,0.55221)
		rgb=(0.21240,0.35968,0.55171)
		rgb=(0.21430,0.35562,0.55118)
		rgb=(0.21621,0.35153,0.55063)
		rgb=(0.21813,0.34743,0.55004)
		rgb=(0.22006,0.34331,0.54941)
		rgb=(0.22199,0.33916,0.54875)
		rgb=(0.22393,0.33499,0.54805)
		rgb=(0.22586,0.33081,0.54731)
		rgb=(0.22780,0.32659,0.54653)
		rgb=(0.22974,0.32236,0.54571)
		rgb=(0.23167,0.31811,0.54483)
		rgb=(0.23360,0.31383,0.54391)
		rgb=(0.23553,0.30953,0.54294)
		rgb=(0.23744,0.30520,0.54192)
		rgb=(0.23935,0.30085,0.54084)
		rgb=(0.24124,0.29648,0.53971)
		rgb=(0.24311,0.29209,0.53852)
		rgb=(0.24497,0.28768,0.53726)
		rgb=(0.24681,0.28324,0.53594)
		rgb=(0.24863,0.27877,0.53456)
		rgb=(0.25043,0.27429,0.53310)
		rgb=(0.25219,0.26978,0.53158)
		rgb=(0.25394,0.26525,0.52998)
		rgb=(0.25565,0.26070,0.52831)
		rgb=(0.25732,0.25613,0.52656)
		rgb=(0.25897,0.25154,0.52474)
		rgb=(0.26057,0.24692,0.52283)
		rgb=(0.26214,0.24229,0.52084)
		rgb=(0.26366,0.23763,0.51876)
		rgb=(0.26515,0.23296,0.51660)
		rgb=(0.26658,0.22826,0.51435)
		rgb=(0.26797,0.22355,0.51201)
		rgb=(0.26931,0.21882,0.50958)
		rgb=(0.27059,0.21407,0.50705)
		rgb=(0.27183,0.20930,0.50443)
		rgb=(0.27301,0.20452,0.50172)
		rgb=(0.27413,0.19972,0.49891)
		rgb=(0.27519,0.19490,0.49600)
		rgb=(0.27619,0.19007,0.49300)
		rgb=(0.27713,0.18523,0.48990)
		rgb=(0.27801,0.18037,0.48670)
		rgb=(0.27883,0.17549,0.48340)
		rgb=(0.27957,0.17060,0.48000)
		rgb=(0.28025,0.16569,0.47650)
		rgb=(0.28087,0.16077,0.47290)
		rgb=(0.28141,0.15583,0.46920)
		rgb=(0.28189,0.15088,0.46541)
		rgb=(0.28229,0.14591,0.46151)
		rgb=(0.28262,0.14093,0.45752)
		rgb=(0.28288,0.13592,0.45343)
		rgb=(0.28307,0.13090,0.44924)
		rgb=(0.28319,0.12585,0.44496)
		rgb=(0.28323,0.12078,0.44058)
		rgb=(0.28320,0.11568,0.43611)
		rgb=(0.28309,0.11055,0.43155)
		rgb=(0.28291,0.10539,0.42690)
		rgb=(0.28266,0.10020,0.42216)
		rgb=(0.28233,0.09495,0.41733)
		rgb=(0.28192,0.08967,0.41241)
		rgb=(0.28145,0.08432,0.40741)
		rgb=(0.28089,0.07891,0.40233)
		rgb=(0.28027,0.07342,0.39716)
		rgb=(0.27957,0.06784,0.39192)
		rgb=(0.27879,0.06214,0.38659)
		rgb=(0.27794,0.05632,0.38119)
		rgb=(0.27702,0.05034,0.37572)
		rgb=(0.27602,0.04417,0.37016)
		rgb=(0.27495,0.03775,0.36454)
		rgb=(0.27381,0.03150,0.35885)
		rgb=(0.27259,0.02556,0.35309)
		rgb=(0.27131,0.01994,0.34727)
		rgb=(0.26994,0.01463,0.34138)
		rgb=(0.26851,0.00961,0.33543)
		rgb=(0.26700,0.00487,0.32942)
	  }
}
\pgfplotscreateplotcyclelist{pastelcolors}{%
  solid, pastelPurple, mark=none\\%
  solid, pastelBlue, mark=none\\%
  solid, pastelGreen, mark=none\\%
  solid, pastelRed, mark=none\\%
  solid, pastelMagenta, mark=none\\%
  solid, pastelOrange, mark=none\\%
  solid, pastelSeaGreen, mark=none\\%
}

\usetikzlibrary{hobby}

\usetikzlibrary{positioning,arrows}
\usepgfplotslibrary{fillbetween}

\begin{document}
\title{Verifying Inverse Model Neural Networks}

\author{Chelsea Sidrane 
\and 
Sydney Katz
\and 
Anthony Corso
\and
Mykel J. Kochenderfer
}

\authorrunning{C. Sidrane et al.}

\institute{Stanford University, Stanford CA 94305, USA \\
\email{\{csidrane, smkatz, acorso, mykel\}@stanford.edu}}

\maketitle

\begin{abstract}
Inverse problems exist in a wide variety of physical domains from aerospace engineering to medical imaging.
The goal is to infer the underlying state from a set of observations. 
When the forward model that produced the observations is nonlinear and stochastic, solving the inverse problem is very challenging. 
Neural networks are an appealing solution for solving inverse problems as they can be trained from noisy data and once trained are computationally efficient to run. 
However, inverse model neural networks do not have guarantees of correctness built-in, which makes them unreliable for use in safety and accuracy-critical contexts. 
In this work we introduce a method for verifying the correctness of inverse model neural networks. 
Our approach is to overapproximate a nonlinear, stochastic forward model with piecewise linear constraints and encode both the overapproximate forward model and the neural network inverse model as a mixed-integer program.
We demonstrate this verification procedure on a real-world airplane fuel gauge case study. 
The ability to verify and consequently trust inverse model neural networks allows their use in a wide variety of contexts, from aerospace to medicine. 

\end{abstract}

\section{Introduction}
Neural networks have been used to solve a variety of nonlinear inverse problems such as state estimation \cite{melzi2011vehicle}, inverse computational mechanics~\citep{TAMADDONJAHROMI2020113217}, inverse-model controller design~\citep{hussain2001implementation}, medical imaging~\citep{hamilton2018deep}, and seismic reflectivity estimation~\citep{kim2018geophysical}. 
Neural networks are used because they can learn complex functions from data and compute outputs quickly, whereas an analytical inverse modeling approach may not work for nonlinear systems, and a sampling-based approach may be computationally expensive.
However, neural network inverse models lack accuracy guarantees, limiting their use in safety critical applications such as transportation. 
To enable the adoption of these models, we present a method for verifying the accuracy of an inverse model neural network. 
Our approach uses recent advancements in neural network verification to provide formal guarantees of the correctness of the model over its entire domain.

Inverse problems arise when trying to estimate an unobserved state from a set of observations, where only the forward model is known (i.e., the model that maps states to observations). 
This is in contrast to a filtering setting where a dynamics model as well as an observation model is often available.
Analytical solutions to inverse problems may not exist for nonlinear systems, and may become intractable when the forward model is high-dimensional or stochastic.
Sampling-based approximate inverse models may be slow to query and may not converge to a useable point estimate at all. 
As a result, data-driven approaches may be applied. 
The forward model is used to produce a dataset of state-observation pairs that are then used to train a neural network to output a state for a given observation.

Once a neural network inverse model is trained, we want guarantees that the inverse mapping between observations and states has low error, but this may be challenging due to the complexity of the forward model and neural network. 
Prior work~\citep{zakrzewski2004randomized} has approached this problem using sampling to get stochastic bounds on the error of an inverse model network. 
Formal approaches~\citep{zakrzewski2002verification} have also been applied to 1-layer networks where the forward model is replaced with a lookup table so that the Lipschitz constants of the simple network and forward model may be easily calculated and the accuracy evaluated at grid points in the domain. 
In addition to these simplifying assumptions, this approach scales exponentially with the number of dimensions in the domain, making it unsuitable for large-scale problems.

We verify ReLU-based inverse model neural networks using mixed-integer linear programming, a technique that has seen significant recent usage for neural network verification~\citep{akintunde2018reachability, tjeng2017evaluating,
dutta2018output, fischetti2017deep, cheng2017maximum, lomuscio2017approach}. 
This approach allows formal verification over the entire input domain without a dependence on gridding.
Instead of replacing the complex forward model that maps states to observations with a lookup table, we encode it into the mixed-integer linear program, preserving the integrity of the model.
This is enabled by overapproximating the nonlinear functions in the forward model using a technique adapted from verification of neural network control systems~\citep{sidrane2021overt}.
The nonlinear functions are overapproximated using piecewise linear constraints, which can be encoded into a mixed-integer linear program, alongside the inverse model neural network.
We then maximize the error between the portion of the state that the inverse model reconstructs and the original state values.
This produces a verified upper bound on the estimation error of the inverse model. \\
\paragraph{Contributions}
Our contribution is to provide a new formal approach to verifying inverse model neural networks by bringing ideas across discipline boundaries -- from reachability to inverse models.
Compared to prior work, our work:
\begin{compactenum}
    \item Can handle multi-layer ReLU neural networks.
    \item Does not require gridding the state space and incurring the curse of dimensionality.
    \item Can compare an inverse model neural network to the true nonlinear forward model, rather than a lookup table.
    \item Can verify stochastic forward models with additive noise.
    \item Employs intelligent parallelization in order to reduce verification time.
    \item Demonstrates the approach on a real-world case study of aircraft fuel measurement.
\end{compactenum}

\section{Related Work}
Neural networks are typically evaluated on a test set after training~\citep{Goodfellow-et-al-2016}, so it is natural to consider sampling-based approaches for neural network validation. 
Sampling-based statistical approaches to validation~\citep{Corso2021survey} have been used for a wide variety of applications such as autonomous driving~\citep{huang2018versatile} and aviation~\citep{zakrzewski2004randomized}. 
However, \citet{koopman2016challenges} argue that if a failure rate of one failure per $10^9$ hours is desired, which is similar to aviation requirements, then at least $10^9$ hours (more than 100,000 years) of testing is required.
Similarly, \citet{zakrzewski2004randomized} investigates the number of samples required to estimate the failure probability with high confidence, considering Bayesian and non-Bayesian settings. 
\citet{zakrzewski2004randomized} ascertains that the number of required samples, on the order of $10^{10}$, is tractable if the system is fast enough to evaluate.
However, in the case of a component for a real autonomous vehicle, where collecting a single sample of a behavior of interest such as merging or turning left could take on the order of minutes, it would not be feasible to collect $10^9$ to $10^{10}$ minutes (1000 - 10,000 years) of driving time.
Additionally, \citet{zakrzewski2004randomized} assumes the ability to sample from the true distribution governing the input domain.
If sampling is performed using a distribution that is slightly wrong, the claims of safety assurance fall apart.
Consequently, other methods obtain guarantees using sampling coverage metrics over the input space, which again scale exponentially with the dimension of the input domain~\citep{Corso2021survey}.
Given the difficulties of sampling-based safety guarantees, we believe formal guarantees which hold over the entire input domain are more compelling.

There has recently been a surge in development of formal methods to verify neural networks~\citep{liu2020algorithms}.
Of particular relevance are approaches which employ mixed-integer programming~\citep{akintunde2018reachability, tjeng2017evaluating,
dutta2018output, fischetti2017deep, cheng2017maximum, lomuscio2017approach}. 
While these approaches can make formal claims about neural networks, they are generally limited to verifying the network over a well-defined input domain. 
In our setting, however, the input domain of the neural network is implicitly defined by a nonlinear, stochastic forward model.

\citet{zakrzewski2002verification} verifies inverse models consisting of 1-layer sigmoid neural networks by replacing the forward model with a lookup table and gridding the input domain.
Our work builds upon this by verifying multi-layer ReLU neural networks, employing a mixed-integer programming approach that does not require gridding the input space, and verifying the inverse model neural network against the true nonlinear model instead of an approximate lookup table.
\citet{zakrzewski2002verification} identifies the difficulty of representing the possible observation space given a domain in state space and a complex forward model.
We represent this set implicitly by overapproximating and encoding the nonlinear measurement model
using an overapproximation technique introduced by \citet{sidrane2021overt}.

\section{Problem Definition}
\label{sec:prob_def}
The state of the system is a vector of variables $\vect{x}$.
Let $\vect{y}$ be a noisy measurement of the state such that
\begin{equation}
    \vect{y} = f(\vect{x},\vects{\nu})
\end{equation}
where $\vects{\nu}$ is a vector of random variables and $f$ is the stochastic forward model.
We are interested in estimating a subset of the state, $\vect{z}$, from a noisy measurement $\vect{y}$.
The function $f$ leads to the distribution $p(\vect{y} \mid \vect{x})$, from which we can write
\begin{equation} \label{eq:marg}
    p(\vect{y} \mid \vect{z}) = \int_{\vect{w}}p(\vect{y} \mid \vect{z}, \vect{w})~p(\vect{w})~d\vect{w}
\end{equation}
where $\vect{x} = [\vect{z}, \vect{w}]^T$.
Our goal is to invert $p(\vect{y} \mid \vect{z})$ and estimate $\vect{z}$ given measurements $\vect{y}$:
\begin{equation}\label{eq:argmax}
    \vect{z} = {\arg \max}_{\vect{z'}} p(\vect{z'} \mid \vect{y})
\end{equation}
without explicitly computing \cref{eq:marg,eq:argmax}.
We interchangeably use distributional notation such as $\vect{y} \sim p(\vect{y} \mid \vect{x})$ and functional notation such as $\vect{y} = f(\vect{x}, \vects{\nu})$ depending on which is more clear for the context.
For the purpose of this paper we define
\begin{equation}
    f^{-1}(\vect{y}) = {\arg \max}_{\vect{z'}} p(\vect{z'} \mid \vect{y})
\end{equation}
The functional inverse may not always exist but ${\arg \max}_{\vect{z'}} p(\vect{z'} \mid \vect{y})$ always exists.

\begin{figure}[t]
\centering
\begin{subfigure}[c]{0.3\textwidth}
    \centering
    \begin{tikzpicture}
    
    \draw (0,0) rectangle (3.2,3.2);
    \fill[pastelBlue!40!white] (0.2,0) rectangle (3.0, 3.2);
    
    \fill[black] (0,0) circle (0.1cm);
    \fill[black] (3.2,0) circle (0.1cm);
    
    \node[anchor=north] at (0.0, 0.0) {\scriptsize Station 1};
    \node[anchor=north] at (3.2, 0.0) {\scriptsize Station 2};
    
    \node[anchor=north] at (1.6, 0.0) {\scriptsize $x$};
    \node[anchor=east] at (0.0, 1.6) {\scriptsize $y$};
    
    \node[anchor=east] at (0.0, 0.15) {\scriptsize ($x_1$,$y_1$)};
    \node[anchor=west] at (3.2, 0.15) {\scriptsize ($x_2$, $y_2$)};
    
    \fill[black] (2.0, 1.2) ellipse (0.08cm and 0.15cm);
    \fill[black] (1.9, 1.2) ellipse (0.04cm and 0.1cm);
    \fill[black] (2.1, 1.2) ellipse (0.04cm and 0.1cm);
    
    \node[anchor=south] at (2.0, 1.3) {\scriptsize Robot};
    
    \draw (0, 0) -- (2.0, 1.2);
    \draw (3.2, 0) -- (2.0, 1.2);
    
    \node[anchor=south] at (1.0, 0.6) {\scriptsize $r_1$};
    \node[anchor=south] at (2.6, 0.6) {\scriptsize $r_2$};
    
    \node[anchor=east] at (1.95, 1.2) {\scriptsize ($x_r$, $y_r$)};
 
\end{tikzpicture}
    \caption{Schematic of example 2D localization problem.}
    \label{fig:toy_diagram}
\end{subfigure}
\hfill 
\begin{subfigure}[c]{0.6\textwidth}
    \centering
    \begin{tikzpicture}
    
    \path[draw,use Hobby shortcut,closed=true] (0,0) .. (.05,0.33) .. (0.33,1) .. (.1,1.12) .. (-0.26, 0.6) .. (-0.33,0.66) .. (-0.33,.166);
    \draw[thick, pastelBlue] (-0.1, -0.1) -- (0.25, 0.15) -- (0.45, 1.0) -- (0.15, 1.3) -- (-0.7, 0.5) -- cycle;
    \node at (0, -0.5) {\scriptsize measurement};
    
    
    \draw (-3.62, 1) -- (-2.72, 1) -- (-2.72, 0.05) -- (-3.62, 0.05) -- cycle;
    \node at (-3.17, 0.52) {$\mathcal{X}$};
    \node at (-3.17, -0.5) {\scriptsize state};
    
    \draw[->] (-2.72, 0.8) -- (0, 0.8);
    \node[anchor=south] at (-1.5, 0.8) {\scriptsize $f(\vect{x}, \vects{\nu})$};
    \draw[->, pastelBlue] (-2.72, 0.25) -- (-0.52, 0.25);
    \node[anchor=south, pastelBlue] at (-1.5, 0.25) {\scriptsize $\hat{f}(\vect{x}, \vects{\nu})$};
    
    \draw[->] (0.3, 0.8) -- (3.4, 0.8);
    \node[anchor=south] at (1.55, 0.8) {\scriptsize $NN(\vect{y})$};
    \draw[->, pastelBlue] (0.3, 0.25) -- (2.7, 0.25);
    \node[anchor=south, pastelBlue] at (1.55, 0.25) {\scriptsize $NN(\vect{y})$};
    
    \path[draw,use Hobby shortcut,closed=true] (3.3,0) .. (3.63,1) .. (3.4,1.12) .. (3.5, 0.6) .. (2.96,0.66) .. (2.96,.166);
    \draw[thick, pastelBlue] (2.7, -0.1) -- (3.9, -0.1) -- (3.9, 1.2) -- (2.7, 1.2) -- cycle;
    \node at (3.35, -0.5) {\scriptsize state estimate};
\end{tikzpicture}
    \caption{Overview of the method.}
    \label{fig:flowchart}
\end{subfigure}
\caption{Overview of example problem and overall method.}
\end{figure}
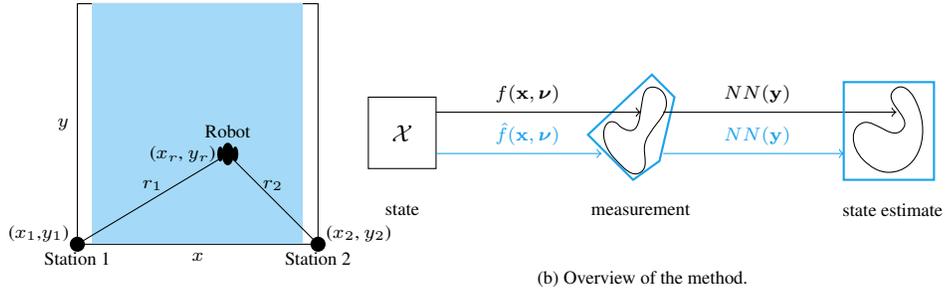

To make the problem more concrete, we will use a 2D localization problem as a running example throughout this section to explain our methodology.  
In this problem, a robot attempts to estimate its position within a 2D room.
The state is the robot's position, $\vect{x} = (x_r,y_r)$ and the robot receives two noisy range measurements of its state:
\begin{equation} \label{eq:nav_mm}
\vect{y} = 
\begin{bmatrix}
    r_1 \\ r_2
\end{bmatrix}
= 
\begin{bmatrix}
    \sqrt{(x_r - x_1)^2 + (y_r - y_1)^2} + \nu_1 \\
    \sqrt{(x_r - x_2)^2 + (y_r - y_2)^2} + \nu_2
\end{bmatrix}
\end{equation}
from sensor beacons located at the corners of the back wall, as shown in \cref{fig:toy_diagram}. 
The sensor beacons are located at coordinates $(x_1, y_1)$ and $(x_2, y_2)$ and each measurement has additive noise $\nu_i \sim \mathcal{N}(0, \sigma)$. 
The robot is free to move anywhere in the blue shaded region.
We would like to  invert the forward model in \cref{eq:nav_mm} and recover the robot's position. In this example, $\vect{z}$ is the same as $\vect{x}$:
\begin{equation}
    \vect{z} = \vect{x} = (x_r,y_r)
\end{equation}

We verify a neural network $NN(\vect{y})$ trained to estimate $\vect{z}$ from a measurement $\vect{y}$ and produce an estimate $\hat{\vect{z}}$:
\begin{equation}
    \hat{\vect{z}} = NN(\vect{y}) \approx {\arg \max}_{\vect{z'}} p(\vect{z'} \mid \vect{y})
\end{equation}
For the localization example, a neural network is trained to produce estimates of the current position $\hat{\vect{z}} = [\hat{x}_r, \hat{y}_r]$ given range measurements $\vect{y} = [r_1, r_2]$ as inputs and true position $\vect{z} = [x_r,y_r]$ as labels.
Once trained, we would like assess the quality of our inverse model. We would like to bound the error between $\vect{z}$ and our estimate of it $\hat{\vect{z}}$. 
Given a bounded region of states $\vect{x} \in \mathcal{X}$, which in turn bounds $\vect{z}$ as $\vect{z} \subseteq \vect{x}$, 
 we want to find an upper bound $u$ such that
 \begin{equation}
     \max_{\vect{x} \in \mathcal{X}} \|\vect{z} - \hat{\vect{z}}\| < u
 \end{equation}
 where $\|\cdot\|$ is some norm.
Neural networks do not have built in guarantees of correctness, so we use neural network verification tools to obtain $u$.
 However, the nonlinear and stochastic nature of the forward model presents new challenges that many existing neural network verification tools are not equipped to handle.
 
\section{Methods} \label{sec:methods}
Our methodology is illustrated in \cref{fig:flowchart}. 
The forward model $f(\vect{x}, \vects{\nu})$ maps states in $\mathcal{X}$ to a set of measurements as shown by the black irregularly shaped set in the center of \cref{fig:flowchart}. 
Our approach is to overapproximate the nonlinear forward model using piecewise linear constraints to produce a model $\hat{f}(\vect{x}, \vects{\nu})$. 
This model implicitly defines an overapproximation of the set of possible measurements which is shown by the blue set in the center of \cref{fig:flowchart}.
Noting that both the overapproximate forward model and the neural network can be represented by piecewise linear expressions, they can be combined into a single mixed-integer linear program. 
We add the maximum error between the reconstruction $\hat{\vect{z}}$ and the original subset of the state $\vect{z}$ as an objective. By solving the resulting optimization problem, we can obtain a bound on the set of possible estimation errors.
The following subsections provide more details.

 
\subsection{Overapproximating the measurement model}\label{sec:OVERT}
The measurement model $f$ which produces measurements $\vect{y}$ may contain polynomials (e.g., $x^2$), roots (e.g., $x^{\frac{1}{2}}$), trigonometric functions (e.g., $\cos(x)$), piecewise nonlinearities (e.g., $\max(x,y)$), exponentials (e.g., $e^x$) and compositions thereof. 
As the end goal is to encode the entire verification problem into a mixed integer \emph{linear} program, we draw on the field of reachability analysis and use the approximation developed by \citet{sidrane2021overt} to overapproximate nonlinear functions with piecewise linear constraints. 
An overapproximation is an approximation such that the set of satisfying values for the overapproximated expression is a superset of the set of values satisfying the original expression.
For example, we would like to quantify  the maximum error between the neural network estimate $\hat{\vect{z}}$ and the true $\vect{z}$.
If we compute an upper bound $\hat{u}$ on the maximum error using the overapproximate model, we know that $\hat{u}$ is greater than or equal to the maximum error $e_{\max}$ that exists between $\vect{z}$ and $\hat{\vect{z}}$ under the true nonlinear model: 
$\hat{u} \geq e_{\max} = \max_{\vect{z}} \|\vect{z} - \hat{\vect{z}}\|$.


 In the example localization problem, the measurement model contains the functions $g(x) = x^2$ and $g(x) = \sqrt{x}$. These functions are converted to piecewise linear bounds to produce the overapproximate forward model $\hat{f}$.
  An illustration of an overapproximation for the $x^2$ function over the domain $[1, 15]$ is shown in \cref{fig:overt_example}.
The lower bound contains 5 linear segments and the upper bound contains 4.
In this example, the error between the function and bound is within 2\% of the magnitude of the range of the function over the specified domain.

\subsection{Posing the verification problem as an MILP} \label{sec:MILP_form}
Once the nonlinear measurement model $f$ has been overapproximated, the entire system consists of piecewise linear relations, because the ReLU-based inverse model neural network $NN(\vect{y})$ is already a piecewise linear function.
Piecewise linear constraints can be encoded into mixed-integer linear programs using binary variables. 
Thus the entire system can be encoded into a mixed integer program.
In order to compute the maximum error between $\vect{z}$ and the estimate returned by the inverse model neural network, $\hat{\vect{z}}$, the following mixed integer optimization problem is solved:
        \begin{maxi}|l|
            {\vect{x}}{|\vect{z} - \hat{\vect{z}}|}{}{}
                    \label{eq:MILP_form}
              \addConstraint{\vect{x}}{\in \mathcal X}{} 
              \addConstraint{\vects{\nu}}{\in N}{}
            \addConstraint{\vect{y}}{\bowtie \hat{f}(\vect{x}, \vects{\nu})}{}
            \addConstraint{\hat{\vect{z}}}{=NN(\vect{y})}{}
        \end{maxi}
where $\mathcal X$ denotes the domain of the state variables $\vect{x}$, of which $\vect{z}$ is a subset, $N$ denotes the domain of the random variables $\vects{\nu}$, and $\vect{y} \bowtie \hat{f}(\vect{x})$ denotes that the measurements $\vect{y}$ are related to the state $\vect{x}$ via the approximated nonlinear measurement model $\hat{f}$. 
The optimal solution to the mixed-integer linear program $(\vect{x}^*, p^*)$ is produced using the commercial MILP solver Gurobi~\citep{gurobi}.
The value of the optimal solution is an upper bound on the maximum error: $p^* = \hat{u} \geq e_{\max}$.
It is an upper bound due to the looseness introduced by overapproximation.

\subsection{Encoding piecewise linear functions into the MILP}\label{sec:max_abs_encodings}
Prior work has demonstrated how ReLU activation functions and other piecewise linear functions may be encoded as mixed-integer constraints~\citep{tjeng2017evaluating,lomuscio2017approach,sidrane2021overt}. 
\citet{tjeng2017evaluating} introduced a tight encoding for $t = \max(x_i), i=1...n$, and $t = \text{ReLU}(x)$ that improves on the classic ``big-M'' encoding~\cite{aps2018mosek} by uniquely bounding the domain of each piecewise linear function.
We apply the same idea to derive a tight encoding of absolute value based on \citep{aps2018mosek} and describe it in \cref{app:tjeng} alongside the other encodings that we use from~\citet{tjeng2017evaluating}.
Necessary for all of these encodings are bounds $[l_i,u_i]$ on each argument $x_i$ to the function.
Bounds $[l_i,u_i]$ are obtained by solving the relaxation of the MILP with $\max x_i$ and then $\min x_i$ as the objective. 
In place of the custom translation from piecewise linear functions to mixed integer constraints that was released by \citet{sidrane2021overt}, we developed a complementary open-source library called \texttt{Expr2MILP.jl} that implements the above encodings. 
It is modular and capable of encoding general piecewise linear expressions composed of $|\cdot|$, $\max$, and $\text{ReLU}$ into mixed-integer constraints. 

\subsection{Handling stochasticity} \label{sec:stochasticity}
Because the forward model is probabilistic, the bounds on the maximum estimation error will also be probabilistic, which we account for using the notion of probability mass. 
When the forward model is overapproximated using the methods in \cref{sec:OVERT}, 
a domain on each variable in the model must be specified. 
Therefore, we restrict the domain of the random variables $\vects{\nu}$ to contain a desired amount of probability mass. 
The probability mass captured by this restricted domain will also correspond to the probability mass captured in the distribution of estimation errors.

For example, the localization problem has two random variables ($\nu_1$ and $\nu_2$), each independently distributed according to a normal distribution with standard deviation $\sigma$.
If we constrain the domain of $\nu_1$ to $\nu_1 \in [-3\sigma$,$3\sigma]$, we capture $99.73\%$ of the probability mass associated with $\nu_1$. 
Similar logic applies to $\nu_2$. 
Because the variables are independent, these domains cover $0.9973^2 = 0.9946$, or $99.46\%$ of the probability mass of the joint distribution. 
In turn, the verification algorithm will find the maximum estimation error across $99.46\%$ of the full distribution of network estimation errors. 
We note that the domains of the random variables can be selected based on the user's desires. 
If we instead constrained the variables to have a domain of $\pm 2\sigma$, we would capture $91.107\%$ of the probability mass in the joint distribution. 
This approach is distinct from a sampling-based approach which can only claim to have captured a certain fraction of the probability mass with a given \emph{confidence level}, whereas we can claim with certainty that we have captured e.g., $99.46\%$ of the probability mass.\footnote{The exact amount of probability mass captured depends on the assumed noise model.}

\section{Example results} \label{sec:nav_results}
\begin{figure}[h!]
\centering 
\begin{subfigure}[b]{0.45\textwidth}
    \centering
    \begin{tikzpicture}[]
    \begin{axis}[
      legend style = {font=\scriptsize, at={(0.1,1.05)}, anchor=south west},
      ylabel = {$g(x)$},
      xlabel = {$x$},
      width=5.2cm, height=5.2cm
    ]
    
    \addplot+[
      solid, black, mark=none
    ] coordinates {
      (1.0, 1.0)
      (1.1414141414141414, 1.3028262422201817)
      (1.2828282828282829, 1.645648403224161)
      (1.4242424242424243, 2.028466483011938)
      (1.5656565656565657, 2.4512804815835123)
      (1.707070707070707, 2.9140903989388836)
      (1.8484848484848484, 3.416896235078053)
      (1.9898989898989898, 3.95969799000102)
      (2.1313131313131315, 4.542495663707785)
      (2.272727272727273, 5.165289256198348)
      (2.414141414141414, 5.8280787674727055)
      (2.5555555555555554, 6.530864197530863)
      (2.696969696969697, 7.273645546372818)
      (2.8383838383838382, 8.056422813998571)
      (2.9797979797979797, 8.87919600040812)
      (3.121212121212121, 9.74196510560147)
      (3.2626262626262625, 10.644730129578614)
      (3.404040404040404, 11.587491072339557)
      (3.5454545454545454, 12.570247933884298)
      (3.686868686868687, 13.593000714212835)
      (3.8282828282828283, 14.655749413325172)
      (3.9696969696969697, 15.758494031221304)
      (4.111111111111111, 16.90123456790123)
      (4.252525252525253, 18.083971023364963)
      (4.393939393939394, 19.306703397612484)
      (4.5353535353535355, 20.569431690643814)
      (4.6767676767676765, 21.87215590245893)
      (4.818181818181818, 23.214876033057852)
      (4.959595959595959, 24.597592082440563)
      (5.101010101010101, 26.020304050607084)
      (5.242424242424242, 27.48301193755739)
      (5.383838383838384, 28.985715743291504)
      (5.525252525252525, 30.528415467809406)
      (5.666666666666667, 32.111111111111114)
      (5.808080808080808, 33.733802673196614)
      (5.94949494949495, 35.396490154065916)
      (6.090909090909091, 37.099173553719005)
      (6.232323232323233, 38.841852872155904)
      (6.373737373737374, 40.62452810937659)
      (6.515151515151516, 42.44719926538109)
      (6.656565656565657, 44.30986634016937)
      (6.797979797979798, 46.21252933374145)
      (6.9393939393939394, 48.15518824609734)
      (7.08080808080808, 50.13784307723701)
      (7.222222222222222, 52.160493827160494)
      (7.363636363636363, 54.22314049586777)
      (7.505050505050505, 56.32578308335884)
      (7.646464646464646, 58.46842158963371)
      (7.787878787878788, 60.65105601469238)
      (7.929292929292929, 62.87368635853484)
      (8.070707070707071, 65.13631262116111)
      (8.212121212121213, 67.43893480257118)
      (8.353535353535353, 69.78155290276501)
      (8.494949494949495, 72.16416692174268)
      (8.636363636363637, 74.58677685950414)
      (8.777777777777779, 77.0493827160494)
      (8.919191919191919, 79.55198449137842)
      (9.06060606060606, 82.09458218549128)
      (9.202020202020202, 84.67717579838792)
      (9.343434343434344, 87.29976533006838)
      (9.484848484848484, 89.96235078053259)
      (9.626262626262626, 92.66493214978064)
      (9.767676767676768, 95.40750943781248)
      (9.909090909090908, 98.19008264462808)
      (10.05050505050505, 101.01265177022752)
      (10.191919191919192, 103.87521681461075)
      (10.333333333333334, 106.77777777777779)
      (10.474747474747474, 109.72033465972858)
      (10.616161616161616, 112.70288746046322)
      (10.757575757575758, 115.72543617998164)
      (10.8989898989899, 118.78798081828387)
      (11.04040404040404, 121.89052137536984)
      (11.181818181818182, 125.03305785123966)
      (11.323232323232324, 128.21559024589328)
      (11.464646464646465, 131.4381185593307)
      (11.606060606060606, 134.70064279155187)
      (11.747474747474747, 138.00316294255688)
      (11.88888888888889, 141.3456790123457)
      (12.030303030303031, 144.7281910009183)
      (12.171717171717171, 148.15069890827465)
      (12.313131313131313, 151.61320273441487)
      (12.454545454545455, 155.11570247933886)
      (12.595959595959595, 158.6581981430466)
      (12.737373737373737, 162.2406897255382)
      (12.878787878787879, 165.8631772268136)
      (13.02020202020202, 169.52566064687278)
      (13.16161616161616, 173.22813998571573)
      (13.303030303030303, 176.9706152433425)
      (13.444444444444445, 180.7530864197531)
      (13.585858585858587, 184.57555351494747)
      (13.727272727272727, 188.4380165289256)
      (13.868686868686869, 192.3404754616876)
      (14.01010101010101, 196.28293031323335)
      (14.151515151515152, 200.26538108356291)
      (14.292929292929292, 204.28782777267625)
      (14.434343434343434, 208.35027038057342)
      (14.575757575757576, 212.45270890725436)
      (14.717171717171718, 216.59514335271913)
      (14.858585858585858, 220.77757371696765)
      (15.0, 225.0)
    };
    \addlegendentry{{}{$x ^ 2$}}
    
    \addplot+[
      solid, pastelRed, mark=none
    ] coordinates {
      (1.0, 0.776)
      (2.7499999999677134, 4.275999999811203)
      (6.249999999911303, 35.77599999863386)
      (9.749999999982487, 91.77599999966678)
      (13.250000000082922, 172.27600000180337)
      (15.0, 224.776)
    };
    \addlegendentry{{}{lower bound, $n=5$}}
    
    \addplot+[
      solid, pastelBlue, mark=none
    ] coordinates {
      (1.0, 1.224)
      (4.5, 20.474)
      (8.0, 64.224)
      (11.5, 132.474)
      (15.0, 225.224)
    };
    \addlegendentry{{}{upper bound, $n=4$}}
    
    \end{axis}
    \end{tikzpicture}
    \caption{Optimally tight overapproximation bounds of the function $g(x) = x^2$ using $n$ segments per bound. 
    }
    \label{fig:overt_example}
    \end{subfigure}
    \hfill
    \begin{subfigure}[b]{0.45\textwidth}
    \centering
    \input{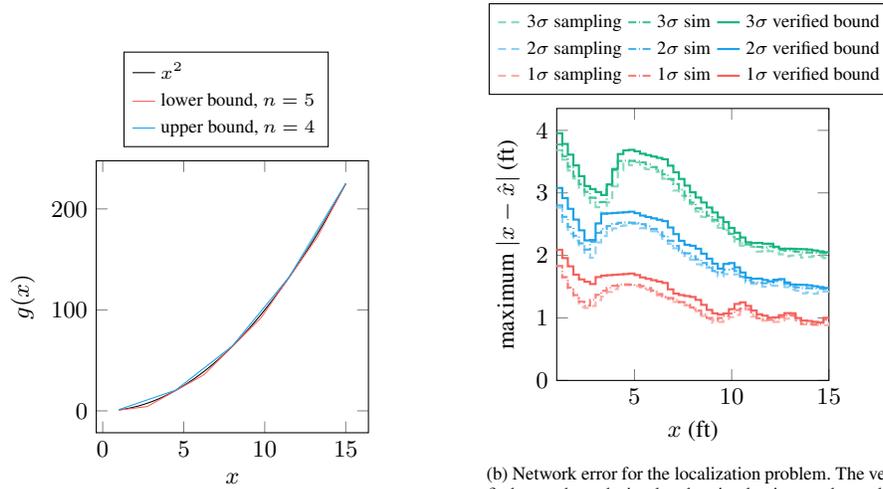}
    \caption{Network error for the localization problem. The verified upper bound, simulated optimal points, and sampled points are close together for the 2D localization problem.}
    \label{fig:nav_err_results}    
\end{subfigure}

\vspace{2ex}

\begin{subfigure}[b]{\textwidth}
    \centering
    \input{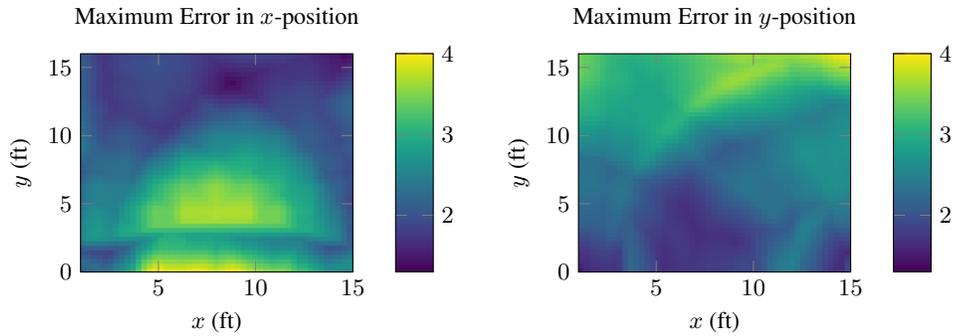}
    \caption{We can quantify the estimation error of our inverse model neural network for the 2D localization problem. Here shown using $3\sigma$ noise levels. We are most certain about the robot's position in the dark blue areas.}
    \label{fig:nav_results}
\end{subfigure}
\caption{Analysis of the 2D localization problem.}
\end{figure}

We apply the methodology described in \cref{sec:methods} to the running example problem of 2D localization.
The first problem maximizes the objective $|x_r - \hat{x}_r|$ where $\hat{x}_r$ is the estimated $x$-position produced by the neural network and $|\cdot|$ is the absolute value.
The second problem maximizes $|y_r - \hat{y}_r|$. 
The choice of norm for the objective is up to the user. For example, $\|\vect{z} - \hat{\vect{z}}\|_{\infty} = \max(|x_r - \hat{x}_r|, |y_r - \hat{y}_r|)$ could have also been chosen.
The problem of finding the maximum error in the $x$ and $y$ directions over the whole domain is solved with two calls to the optimizer, resulting in maximum $x$ and $y$ estimation errors over the whole domain of 3.95\,ft and 3.96\,ft, respectively, for $3\sigma$ noise levels on the range measurements.
In order to better visualize the error, we discretize the domain according to our desired computational budget and solve two optimization problems for each $(x,y)$ cell.
The results can be seen in \cref{fig:nav_results}.
We can tell from these bounds that errors in the estimate of $x$-position tend to be higher for small values of $y$, and errors in the estimate of the $y$-position tends to be higher for large values of $y$. Based on the sensor beacon locations, small changes in range result in larger changes in $x$ for small values of $y$ producing a more difficult estimation problem in that region. The opposite is true for the $y$-position.
A possible conclusion of this analysis might be that additional sensor beacons are needed in the upper corners of the room (see \cref{fig:toy_diagram}), or that more training data is needed from these more difficult regions of state space.

Additionally, for the $x$-coordinate, we compare the upper bounds on error computed using our verification approach to approximate upper bounds obtained through sampling and through the simulation of the extreme points found from verification (labeled ``sim'').
The results are shown in \cref{fig:nav_err_results}. 
The output of the formal verification upper bounds the values from the true nonlinear model.
We could have displayed a single upper bound on the x-position error but in order to better visualize the trends in network error, have discretized the state space and plotted stepwise maximum error curves.


\section{Fuel Gauge Case Study}
      \begin{figure*}[h!]
            \centering
            \input{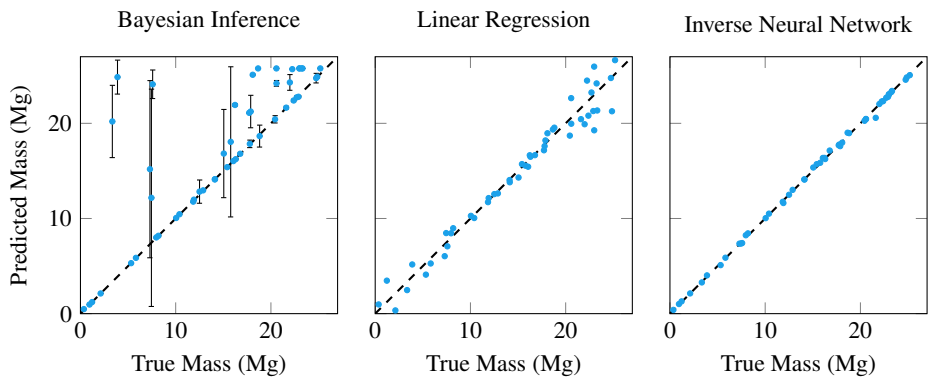}
            \caption{A neural network performs more precisely than a linear regressor and a sampling-based Bayesian method. The error bars in the Bayesian inference plot represent two standard deviations.}
            \label{fig:justify_nn}
        \end{figure*}
We use an airplane fuel gauge as a case study to demonstrate the real-world applicability of our method.
Fuel is often kept in wings of aircraft and getting an accurate measurement 
during flight is difficult. 
As fuel is flight critical, an accurate, responsive, and independent measurement is desired. 
Filtering using a time-history of measurements is undesirable because it allows errors in fuel estimates to propagate across time.
For this reason, single-shot Bayesian inference, which can be used to estimate the distribution $p(\vect{z} \mid \vect{y})$ given the forward model $p(\vect{y} \mid \vect{z})$, is a desirable technique to solve the problem. Nevertheless, exact Bayesian inference is often intractable for systems with nonlinear measurement models. Therefore, these techniques often rely on complex sampling methods such as Markov Chain Monte Carlo (MCMC), which can require many samples to produce accurate estimates and only provide statistical guarantees of convergence \citep{brooks2011handbook}.
All of these properties make the use of an inverse model neural network very appealing as it can produce accurate, responsive and independent estimates. 
        
Nonetheless, we compare the performance of a neural network, linear regression, and sampling-based Bayesian inference for inverting the fuel model.
We use a forward model mapping fuel mass to pressure measurements similar to the model used in \citet{zakrzewski2001fuel}, and then train a ReLU-based inverse model neural network to reconstruct a fuel mass estimate.  
The forward model we use is described in \cref{sec:fuel_gauge_model}.
For the inverse model neural network, we use a fully connected network with three hidden layers containing $64$, $32$, and $12$ hidden units respectively.
We compare to a Bayesian inference technique  called No-U-Turn sampling (NUTS) \citep{hoffman2014no} implemented in \texttt{Turing.jl} \citep{ge2018t} that is a version of MCMC. The NUTS algorithm was used to generate $1000$ samples for each mass value with $1000$ adaptation steps and a target accept rate of $0.9$. A comparison of these techniques is summarized in \cref{fig:justify_nn}. 
Overall, the neural network has the best performance. While the Bayesian inference estimates often converge close to the true value, the MCMC sampling procedure sometimes fails to converge to an accurate estimate. Furthermore, the Bayesian inference queries took $26$ seconds on average on a single 4.20 GHz Intel Core i7 processor, while the neural network queries took \num{6e-6} seconds on average.

However, using a neural network fuel gauge raises safety concerns.
If the neural network produces an inaccurate measurement of remaining fuel, the remaining range of the aircraft could be shorter than the pilot expects. 
To ensure safety, the Federal Aviation Administration has strict certification requirements for all flight software that have largely prevented the use of neural networks on aircraft because there weren't good ways to verify their performance~\citep{nla.cat-vn4510326}.
To bridge this gap, we propose to verify the neural network fuel gauge offline. 
The results of verification could then be used to determine if the network satisfies the requirements for onboard use, and if not, be used to inform system redesign or network retraining. 

\subsection{Airplane fuel gauge model}
The forward model for the airplane fuel gauge mapping fuel mass to pressure measurements is noisy and highly nonlinear, containing piecewise linear functions, multiplication of variables, $\sin$, $\cos$, and polynomials. 
The model is described in more detail in \cref{sec:fuel_gauge_model}.
The state $\vect{x}$ consists of the current mass of fuel in the tank $m$, roll $\phi$ and pitch $\theta$ of the aircraft, and ambient air pressure $P_a$. The forward measurement model computes readings at 9 pressure sensors in the fuel tank as well as the three-dimensional acceleration of the aircraft, which all have additive noise.
The problem is stated as follows:
\begin{align}
    \vect{x} &= [m,~\phi,~\theta,~P_a] \\
    \vects{\nu} &= [\nu_{p_1}, ..., \nu_{p_9}, \nu_{a_x}, \nu_{a_y}, \nu_{a_z}] \\
    \vect{y} &= [p_1, ..., p_9, a_x, a_y, a_z] \\
    \vect{z} &= [m] 
\end{align}
The inverse model attempts to recover the fuel mass $m$, and we want to quantify the error between the neural network estimate of the fuel mass $\hat m$ and its true value, where $|\cdot|$ denotes absolute value: 
\begin{equation}
\max_{\vect{x}}|m-\hat{m}|
\end{equation}

\subsection{Making Verification Tractable}
The fuel gauge model we adapted from \citet{zakrzewski2001fuel} involves a high-resolution lookup table that threatens the tractability of the verification problem.
Despite abundant nonlinearity and the use of transcendental functions such as in \cref{eq:g}, the resulting forward model of the airplane fuel gauge could be encoded and solved as a single mixed integer program \emph{except} for the fact that the analytical mappings from volume, pitch and roll to fuel height at a given pressure sensor, $h_\text{fuel} = f(v, \theta, \phi)$ are not known. 
As a result, lookup tables mapping $(v,\theta, \phi) \rightarrow h_\text{fuel}$  are calculated from a CAD model of the fuel tank and these tables are treated as ground truth.
Luckily, even though a large number of grid cells ($7488$) are imposed on the problem, they are completely independent of one another, and we exploit this by parallelizing across each cell. 

A naive parallelization would not be computationally tractable due to the slow speed of OVERT and the encoding process, but we re-use the overapproximation and encoding in order to regain tractability.
A domain is required for each variable for overapproximation using OVERT as well as for the mixed-integer encoding process for both the network and the overapproximate forward model.
However, if we overapproximated and encoded the MILP separately for the domain of each cell, the computation would take days to finish, because the overapproximation and encoding together requires several minutes for a single cell. 
Our solution is to overapproximate and encode the fuel gauge measurement model and the inverse model neural network over the domain corresponding to the union of all of the cells.
We then duplicate the optimization problem and additionally constrain the domain of each cell, $\mathcal X$ in \cref{eq:MILP_form}, and the mappings from volume, pitch and roll to fuel height, to find the maximum error unique to that specific cell. 
    
Another key modification that was made to increase the speed of verification was the use of a sampled lower bound. 
When seeking the maximum value of $|m - \hat m|$, all samples of the true nonlinear model form a lower bound. 
A constraint is added to the optimization problem for each cell of the form 
        \begin{equation}
            |m - \hat m| > s_{\max}
        \end{equation}
where $s_{\max}$ is the worst deviation between $m$ and $\hat m$ found while sampling from that cell's domain uniformly at random. 
Sampling is fast and helps narrow the search space for an optimum. Overall, we observed a speedup of up to 16\%. 
\section{Fuel Gauge Results}
\begin{figure}[h!]
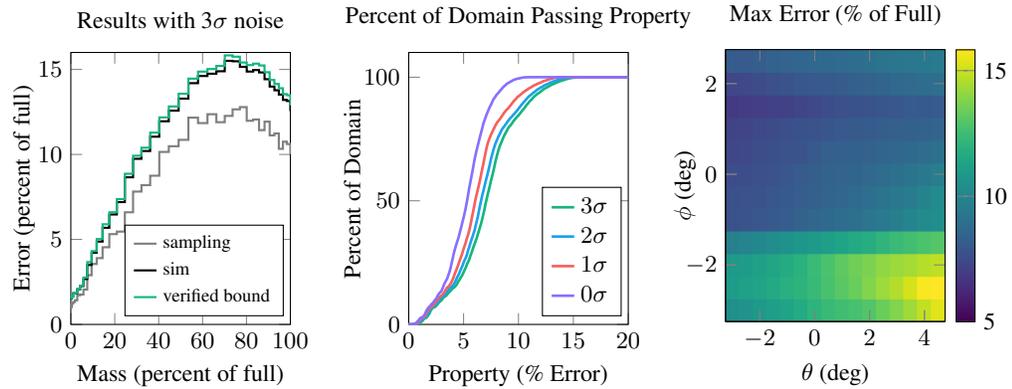

    \centering
    \begin{subfigure}[b]{0.28\textwidth}
        \centering
        \begin{tikzpicture}[]
\begin{groupplot}[group style={horizontal sep = 0.25cm, group size=3 by 1}]

\nextgroupplot [
  height = {5.2cm},
  legend pos = {south east},
  title = {Results with 3$\sigma$ noise},
  xmin = {0.0},
  xmax = {100.0},
  ymax = {16.0},
  xlabel = {Mass (percent of full)},
  ylabel = {Error (percent of full)},
  ymin = {0.0},
  width = {4.5cm},
  legend style={font=\scriptsize},
  ylabel shift = {-5pt},
]

\addplot+[
  mark = {none},
  solid, thick, gray, const plot
] coordinates {
  (0.0, 1.2377736351687272)
  (0.021598272138228937, 1.205552324064028)
  (0.1511879049676026, 1.0224967453194167)
  (0.4751619870410366, 1.2205197024694308)
  (1.0583153347732182, 1.3234468218077682)
  (1.8790496760259179, 1.4230582014079938)
  (2.9373650107991356, 1.7164535923712785)
  (4.233261339092873, 1.7592451587630749)
  (5.745140388768898, 2.052055060720611)
  (7.473002159827215, 2.895116640412936)
  (9.460043196544277, 3.5557360405186116)
  (11.771058315334773, 3.900923023720594)
  (14.449244060475161, 4.539829957941494)
  (17.516198704103672, 5.313563301127395)
  (20.95032397408207, 5.461671870463833)
  (24.600431965442763, 6.59797841921513)
  (28.33693304535637, 7.816834633129437)
  (32.159827213822894, 8.19544344583829)
  (36.13390928725701, 8.74882588998199)
  (40.28077753779697, 10.148941779554224)
  (44.62203023758099, 10.48330532728022)
  (49.09287257019439, 10.859464280798347)
  (53.542116630669554, 12.16508396633109)
  (57.92656587473001, 11.845003082932987)
  (62.224622030237576, 12.366262946103717)
  (66.28509719222461, 11.896024055741439)
  (70.10799136069114, 12.268172482418647)
  (73.67170626349892, 12.65360005315188)
  (76.99784017278618, 12.785703416602209)
  (80.10799136069114, 11.981335543489688)
  (83.02375809935205, 12.058368694668623)
  (85.72354211663067, 11.428832960355983)
  (88.22894168466522, 11.92931838030977)
  (90.56155507559396, 11.122419309937106)
  (92.72138228941685, 11.234566438662567)
  (94.73002159827213, 10.329612444851694)
  (96.60907127429806, 10.751798044463504)
  (98.3585313174946, 10.614485598919252)
  (100.0, 10.014775611363374)
};
\addlegendentry{{}{sampling}}

\addplot+[
  mark = {none},
  solid, thick, black, const plot
] coordinates {
  (0.0, 1.6683155705460468)
  (0.021598272138228937, 1.6469634639649708)
  (0.1511879049676026, 1.5188508244783492)
  (0.4751619870410366, 1.6324361195248092)
  (1.0583153347732182, 1.826717956908109)
  (1.8790496760259179, 1.831065440461515)
  (2.9373650107991356, 2.0692084102842974)
  (4.233261339092873, 2.2580085897869027)
  (5.745140388768898, 2.6699585005361985)
  (7.473002159827215, 3.498089632227518)
  (9.460043196544277, 4.2206268207770306)
  (11.771058315334773, 4.874080612387758)
  (14.449244060475161, 5.6863021157706815)
  (17.516198704103672, 6.468114759316966)
  (20.95032397408207, 7.197364420302983)
  (24.600431965442763, 8.656977298892208)
  (28.33693304535637, 9.743127032084935)
  (32.159827213822894, 10.201588760395179)
  (36.13390928725701, 11.044088807966492)
  (40.28077753779697, 11.9521820575864)
  (44.62203023758099, 12.576619105537059)
  (49.09287257019439, 13.284126927197573)
  (53.542116630669554, 14.19305881179901)
  (57.92656587473001, 14.613966479016247)
  (62.224622030237576, 14.749604917113817)
  (66.28509719222461, 14.894004508174383)
  (70.10799136069114, 15.507604850149793)
  (73.67170626349892, 15.480066221860966)
  (76.99784017278618, 15.153865119421543)
  (80.10799136069114, 14.876053504358373)
  (83.02375809935205, 14.966413266880341)
  (85.72354211663067, 14.825029141573317)
  (88.22894168466522, 14.426102243742928)
  (90.56155507559396, 13.994257291233971)
  (92.72138228941685, 13.775772720618553)
  (94.73002159827213, 13.523831307313186)
  (96.60907127429806, 13.2079049642287)
  (98.3585313174946, 13.145261378491952)
  (100.0, 12.546274743993536)
};
\addlegendentry{{}{sim}}

\addplot+[
  mark = {none},
  thick, pastelSeaGreen, const plot
] coordinates {
  (0.0, 1.6903879027613133)
  (0.021598272138228937, 1.6690357961801956)
  (0.1511879049676026, 1.5409231566936084)
  (0.4751619870410366, 1.6520348997964083)
  (1.0583153347732182, 1.8451502830834507)
  (1.8790496760259179, 1.8494977539559385)
  (2.9373650107991356, 2.0854489549344746)
  (4.233261339092873, 2.2777091289039353)
  (5.745140388768898, 2.7587704172469505)
  (7.473002159827215, 3.668364555139598)
  (9.460043196544277, 4.314642727170688)
  (11.771058315334773, 5.024764194953138)
  (14.449244060475161, 5.892706873237612)
  (17.516198704103672, 6.599209482725262)
  (20.95032397408207, 7.375757107958608)
  (24.600431965442763, 8.864751915822335)
  (28.33693304535637, 9.92551637677102)
  (32.159827213822894, 10.397959372406367)
  (36.13390928725701, 11.2846672672076)
  (40.28077753779697, 12.181678332752858)
  (44.62203023758099, 12.825371495387664)
  (49.09287257019439, 13.57219479009258)
  (53.542116630669554, 14.459290119322882)
  (57.92656587473001, 14.873928968757303)
  (62.224622030237576, 15.02564846874814)
  (66.28509719222461, 15.209701681255872)
  (70.10799136069114, 15.828048067651501)
  (73.67170626349892, 15.750862352383155)
  (76.99784017278618, 15.409122690528978)
  (80.10799136069114, 15.258068094507218)
  (83.02375809935205, 15.342593230098043)
  (85.72354211663067, 15.236970790454137)
  (88.22894168466522, 14.837962843174154)
  (90.56155507559396, 14.382995125401967)
  (92.72138228941685, 14.147224712765288)
  (94.73002159827213, 13.89531325863772)
  (96.60907127429806, 13.551183762571476)
  (98.3585313174946, 13.48854017683494)
  (100.0, 12.88955354233642)
};
\addlegendentry{{}{verified bound}}

\end{groupplot}

\end{tikzpicture}
        \caption{Verified upper bound is tight to the sampled lower bounds.}
        \label{fig:nn_sampling}
    \end{subfigure}
    \hfill
    \begin{subfigure}[b]{0.28\textwidth}
        \centering
        \input{figures/cdf_many_runs}
        \caption{Fuel gauge performance varying noise bounds.}
        \label{fig:cdf}
    \end{subfigure}
    \hfill 
    \begin{subfigure}[b]{0.28\textwidth}
        \centering
        \input{figures/mdiff_2D_pitch_roll_64555}
      \caption{Maximum error of the neural network.}
        \label{fig:roll_pitch_grid}
    \end{subfigure}
    \caption{Analysis of the neural network fuel gauge.}
\end{figure}

We present the results of running our verification method for an inverse model neural network on the airplane fuel gauge case study.
Quantifying the error in fuel mass estimate produced by the inverse model neural network is important because the pilot needs an accurate measurement of how much fuel is left in the plane.
We are able to obtain verified upper bounds $\hat u$ on error $\max_{\vect{x}} |m - \hat{m}|$ 
for each $(v, \theta, \phi)$ cell in the problem domain. 
We maximize over $\phi$ (roll) and $\theta$ (pitch) values in order to find the maximum error at a given fuel level and show the results in \cref{fig:nn_sampling}.
The verification took 1.520 hours or equivalently 0.731 seconds per cell and with 3$\sigma$ noise on the independent variables, and captures $96.81\%$ of the error distribution probability mass. 
Experiments where run on 148 physical cores (see \cref{sec:comp_deets}).

\Cref{fig:nn_sampling} demonstrates both the network performance as well as the tightness of our verified upper bound.
We compare the verified upper bounds on error to two baselines.
The first baseline is uniform random sampling from the domain of $\vect{x}$ and the corresponding noise domain for noise variables $\vects{\nu}$. We estimate the maximum error using $63,000$ samples for each mass bin. 
We can see that the verified upper bound on error is much higher than the sampled error estimate.
Sampling will always form a lower bound to the maximum error, so this result is as expected.
However, the difference between the verified bound and sampling may be due to both overapproximation in the forward model and suboptimality of random sampling.
To determine whether overapproximation in the forward model or suboptimality of random sampling has more impact, we pass the optimum point $\vect{x}^*$ for each cell
through the original nonlinear forward model and the neural network fuel gauge.
The point $\vect{x}^*$ is in effect an adversarial sample found through optimization and produces an error which is plotted under the label ``sim'' in \cref{fig:nn_sampling}. 
This line is tight to the upper bound on error values produced by the optimizer using the overapproximate forward model, and relatively farther from the sampled lower bound.
This result suggests that little conservatism is added from the overapproximation, which is desirable.
It also tells us that the naively sampled lower bound is not as tight to the true maximum error curve (which the ``sim'' curve still lower bounds) as in the example localization problem. 
This problem has many more state variables and 12 dimensions of noise, which make it difficult for unguided random sampling to find the maximum error.
Note that we cannot compare our verification method to the the methods of \citet{zakrzewski2002verification} because their method is not capable of evaluating error with respect to the true nonlinear, stochastic forward model; it is only capable of comparison against a table. 

Our method also allows precise quantification of the error of the inverse model neural network throughout different areas of the state space.
\Cref{fig:roll_pitch_grid} shows the maximum amount of error for different regions of state space as roll and pitch of the aircraft vary. 
The maximum over the entire range of fuel mass values is shown.
The bright yellow spot in the bottom right corner of \cref{fig:roll_pitch_grid} indicating higher error suggests more training data from this region of the state space may be beneficial.
Finally, we propose \cref{fig:cdf} as a metric that regulators could use to evaluate network performance in a more succinct way than \cref{fig:roll_pitch_grid} or \cref{fig:nn_sampling}.
\Cref{fig:cdf} show the percent of the domain within a certain maximum error level, for 4 noise levels.
Simultaneously, \cref{fig:cdf} displays worst case errors ($\sim 15\%)$, illustrates  the distribution of errors,  and shows how the distribution of errors changes as more noise is considered.
For  example,  we can see that most of the cells  have $5-10\%$ error for all noise levels.
Again, this may indicate that network retraining is needed.

\section{Discussion}
\label{sec:discussion}
\textbf{Limitations} While our method is much more scalable than historical methods such as those presented by \citet{zakrzewski2002verification, zakrzewski2004randomized}, it cannot scale to the largest modern networks (1mil+ parameters). 
Additionally, while the fuel measurement model is very complex,
our approach would slow down with more nonlinearity such as additional layers of nested multiplication, e.g. if all constants in the fuel model were allowed to vary as well.
This is due to the fact that multiplication is a costly operation for OVERT to overapproximate and increases the size of the MILP substantially.
Given the fact that this is the first paper to apply MILP-based NNV to inverse models, and that there has been a recent resurgence in NNV research, we expect the performance of our overall approach to improve as NNV and MILP-solving technology continues to improve~\citep{bak2021second, bixby2015computational}. 
Nevertheless, we are still able to tackle a real-world problem: fuel gauging, for which 
a small network worked well (see \cref{fig:justify_nn}).


\vspace{10pt}
\noindent \textbf{Conclusion}
We introduced a method for quantifying the error of inverse model neural networks against nonlinear, stochastic forward models.
Given a noise threshold on each of the random variables in the model, our method produces verified upper bounds on the estimation error.
In addition, the points of the domain selected by the optimizer represent adversarial samples of the true model that may not be found with unguided sampling. 
We demonstrate that our method can scale to real-world problems such as an airplane fuel gauge system. 
In general, neural networks do not have built-in guarantees of correctness, which can prevent their application to many real world problems that could benefit from their flexibility and 
performance.
Precisely quantifying estimation error of inverse model neural networks could allow their use in many safety-critical applications.

\subsubsection{Acknowledgements}
This work was supported by AFRL and DARPA under contract FA8750-18-C-0099.
The NASA University Leadership initiative (grant 80NSSC20M0163) provided funds to assist the authors with their research, but this article solely reflects the opinions and conclusions of its authors and not any NASA entity.
This research was also supported by the National Science Foundation Graduate Research Fellowship under Grant No. DGE–1656518.

\bibliography{bib}

\begin{thebibliography}{28}
\providecommand{\natexlab}[1]{#1}
\providecommand{\url}[1]{\texttt{#1}}
\providecommand{\urlprefix}{URL }
\expandafter\ifx\csname urlstyle\endcsname\relax
  \providecommand{\doi}[1]{doi:\discretionary{}{}{}#1}\else
  \providecommand{\doi}{doi:\discretionary{}{}{}\begingroup
  \urlstyle{rm}\Url}\fi

\bibitem[{Akintunde et~al.(2018)Akintunde, Lomuscio, Maganti, and
  Pirovano}]{akintunde2018reachability}
Akintunde, M., Lomuscio, A., Maganti, L., Pirovano, E.: Reachability analysis
  for neural agent-environment systems. In: International Conference on
  Principles of Knowledge Representation and Reasoning (2018)

\bibitem[{Bak et~al.(2021)Bak, Liu, and Johnson}]{bak2021second}
Bak, S., Liu, C., Johnson, T.: The second international verification of neural
  networks competition (vnn-comp 2021): Summary and results. arXiv preprint
  arXiv:2109.00498  (2021)

\bibitem[{Bixby et~al.(2004)Bixby, Fenelon, Gu, Rothberg, and
  Wunderling}]{bixby2015computational}
Bixby, R.E., Fenelon, M., Gu, Z., Rothberg, E., Wunderling, R.: Mixed-Integer
  Programming: A Progress Report, pp. 309--325 (2004),
  \doi{10.1137/1.9780898718805.ch18}

\bibitem[{Brooks et~al.(2011)Brooks, Gelman, Jones, and
  Meng}]{brooks2011handbook}
Brooks, S., Gelman, A., Jones, G., Meng, X.L.: Handbook of {M}arkov {C}hain
  {M}onte {C}arlo. {CRC} Press (2011)

\bibitem[{Cheng et~al.(2017)Cheng, N{\"u}hrenberg, and
  Ruess}]{cheng2017maximum}
Cheng, C.H., N{\"u}hrenberg, G., Ruess, H.: Maximum resilience of artificial
  neural networks. In: International Symposium on Automated Technology for
  Verification and Analysis, pp. 251--268, Springer (2017)

\bibitem[{Corso et~al.(2021)Corso, Moss, Koren, Lee, and
  Kochenderfer}]{Corso2021survey}
Corso, A., Moss, R.J., Koren, M., Lee, R., Kochenderfer, M.J.: A survey of
  algorithms for black-box safety validation of cyber-physical systems. Journal
  of Artificial Intelligence Research \textbf{72}(2005.02979), 377--428 (2021),
  \doi{10.1613/jair.1.12716}

\bibitem[{Dutta et~al.(2018)Dutta, Jha, Sankaranarayanan, and
  Tiwari}]{dutta2018output}
Dutta, S., Jha, S., Sankaranarayanan, S., Tiwari, A.: Output range analysis for
  deep feedforward neural networks. In: NASA Formal Methods Symposium, pp.
  121--138, Springer (2018)

\bibitem[{Fischetti and Jo(2017)}]{fischetti2017deep}
Fischetti, M., Jo, J.: Deep neural networks as 0-1 mixed integer linear
  programs: A feasibility study. arXiv preprint arXiv:1712.06174  (2017)

\bibitem[{Ge et~al.(2018)Ge, Xu, and Ghahramani}]{ge2018t}
Ge, H., Xu, K., Ghahramani, Z.: Turing: a language for flexible probabilistic
  inference. In: International Conference on Artificial Intelligence and
  Statistics, pp. 1682--1690 (2018),
  \urlprefix\url{http://proceedings.mlr.press/v84/ge18b.html}

\bibitem[{Goodfellow et~al.(2016)Goodfellow, Bengio, and
  Courville}]{Goodfellow-et-al-2016}
Goodfellow, I., Bengio, Y., Courville, A.: Deep Learning. MIT Press (2016),
  \url{http://www.deeplearningbook.org}

\bibitem[{{Gurobi Optimization, LLC}(2022)}]{gurobi}
{Gurobi Optimization, LLC}: {Gurobi Optimizer Reference Manual} (2022),
  \urlprefix\url{https://www.gurobi.com}

\bibitem[{Hamilton and Hauptmann(2018)}]{hamilton2018deep}
Hamilton, S.J., Hauptmann, A.: Deep d-bar: Real-time electrical impedance
  tomography imaging with deep neural networks. IEEE Transactions on Medical
  Imaging \textbf{37}(10), 2367--2377 (2018)

\bibitem[{Hoffman et~al.(2014)Hoffman, Gelman et~al.}]{hoffman2014no}
Hoffman, M.D., Gelman, A., et~al.: The {N}o-{U}-{T}urn sampler: adaptively
  setting path lengths in {H}amiltonian {M}onte {C}arlo. Journal of Machine
  Learning Research \textbf{15}(1), 1593--1623 (2014)

\bibitem[{Huang et~al.(2018)Huang, Guo, Arief, Lam, and
  Zhao}]{huang2018versatile}
Huang, Z., Guo, Y., Arief, M., Lam, H., Zhao, D.: A versatile approach to
  evaluating and testing automated vehicles based on kernel methods. In:
  American Control Conference, pp. 4796--4802, IEEE (2018)

\bibitem[{Hussain et~al.(2001)Hussain, Kittisupakorn, and
  Daosu}]{hussain2001implementation}
Hussain, M.A., Kittisupakorn, P., Daosu, W.: Implementation of
  neural-network-based inverse-model control strategies on an exothermic
  reactor. Science Asia \textbf{27}, 41--50 (2001)

\bibitem[{Kim and Nakata(2018)}]{kim2018geophysical}
Kim, Y., Nakata, N.: Geophysical inversion versus machine learning in inverse
  problems. The Leading Edge \textbf{37}(12), 894--901 (2018)

\bibitem[{Koopman and Wagner(2016)}]{koopman2016challenges}
Koopman, P., Wagner, M.: Challenges in autonomous vehicle testing and
  validation. SAE International Journal of Transportation Safety \textbf{4}(1),
  15--24 (2016)

\bibitem[{Liu et~al.(2021)Liu, Arnon, Lazarus, Strong, Barrett, and
  Kochenderfer}]{liu2020algorithms}
Liu, C., Arnon, T., Lazarus, C., Strong, C., Barrett, C., Kochenderfer, M.J.:
  Algorithms for verifying deep neural networks. Foundations and Trends in
  Optimization \textbf{4}(3--4), 244--404 (2021), \doi{10.1561/2400000035},
  \urlprefix\url{https://arxiv.org/abs/1903.06758}

\bibitem[{Lomuscio and Maganti(2017)}]{lomuscio2017approach}
Lomuscio, A., Maganti, L.: An approach to reachability analysis for
  feed-forward relu neural networks. arXiv preprint arXiv:1706.07351  (2017)

\bibitem[{Melzi and Sabbioni(2011)}]{melzi2011vehicle}
Melzi, S., Sabbioni, E.: On the vehicle sideslip angle estimation through
  neural networks: Numerical and experimental results. Mechanical Systems and
  Signal Processing \textbf{25}(6), 2005--2019 (2011)

\bibitem[{{Mosek ApS}(2018)}]{aps2018mosek}
{Mosek ApS}: {MOSEK Modeling Cookbook} (2018)

\bibitem[{RTCA(1992)}]{nla.cat-vn4510326}
RTCA: RTCA/DO-178B, EUROCAE/ED-12B: Software Considerations in Airborne Systems
  and Equipment Certification (1992)

\bibitem[{Sidrane et~al.(2022)Sidrane, Maleki, Irfan, and
  Kochenderfer}]{sidrane2021overt}
Sidrane, C., Maleki, A., Irfan, A., Kochenderfer, M.J.: {OVERT}: An algorithm
  for safety verification of neural network control policies for nonlinear
  systems. Journal of Machine Learning Research \textbf{23}(117), 1--45 (2022)

\bibitem[{Tamaddon-Jahromi et~al.(2020)Tamaddon-Jahromi, Chakshu, Sazonov,
  Evans, Thomas, and Nithiarasu}]{TAMADDONJAHROMI2020113217}
Tamaddon-Jahromi, H.R., Chakshu, N.K., Sazonov, I., Evans, L.M., Thomas, H.,
  Nithiarasu, P.: Data-driven inverse modelling through neural network (deep
  learning) and computational heat transfer. Computer Methods in Applied
  Mechanics and Engineering \textbf{369}, 113217 (2020)

\bibitem[{Tjeng et~al.(2019)Tjeng, Xiao, and Tedrake}]{tjeng2017evaluating}
Tjeng, V., Xiao, K., Tedrake, R.: Evaluating robustness of neural networks with
  mixed integer programming. In: International Conference on Learning
  Representations (2019)

\bibitem[{Zakrzewski(2001)}]{zakrzewski2001fuel}
Zakrzewski, R.R.: Fuel volume measurement in aircraft using neural networks.
  In: International Joint Conference on Neural Networks, vol.~1, pp. 687--692,
  IEEE (2001)

\bibitem[{Zakrzewski(2002)}]{zakrzewski2002verification}
Zakrzewski, R.R.: Verification of performance of a neural network estimator.
  In: International Joint Conference on Neural Networks, vol.~3, pp.
  2632--2637, IEEE (2002)

\bibitem[{Zakrzewski(2004)}]{zakrzewski2004randomized}
Zakrzewski, R.R.: Randomized approach to verification of neural networks. In:
  International Joint Conference on Neural Networks, vol.~4, pp. 2819--2824,
  IEEE (2004)

\end{thebibliography}


\newpage
\appendix
\onecolumn
\section{Appendix on Piecewise Linear Encodings} \label{app:tjeng}
This appendix describes the encodings of piecewise linear functions into a mixed-integer linear program.
The overapproximation of the forward model contains $\max$ functions.
Using the classic ``big-M'' method ~\citep{aps2018mosek}, the function $t = \max\{x_1,...,x_n\}$ would be encoded
   \begin{subequations} 
    \begin{align}
    x_i &\leq t \leq x_i + M(1-z_i), i=1,...,n, \\
    z_1 + \cdots + z_n &= 1 \\
    z &\in \{0,1\}^n
    \end{align} 
    \end{subequations}
where $z_i$ indicates which variable $x_i$ is the maximum and $M$ is a very large constant upper bounding the difference between the maximum and the other inputs:  $M \geq x_j^* - x_{i \neq j}$. 
Using a large constant $M$ leads to a loose approximation of the output range of the $\max$ function, and as discussed earlier, this can lead to slower solve times.
In contrast, \citet{tjeng2017evaluating} introduces the following encoding which is much tighter:
\begin{subequations} 
    \label{eq:tjeng_max}
    \begin{align}
    x_i &\leq t \leq x_i + (u_{\max_i} - l_i)(1 - z_i) \\
    z_1 + \cdots + z_n &= 1 \\
    z &\in \{0,1\}^n
\end{align} 
\end{subequations}
where each $x_i$ has bounds $[l_i, u_i]$, $u_{\max_i} = \max_{j\neq i} u_j$, and the maximum is only taken over inputs where $u_i \geq l_{\max} =  \max_i l_i$, because if the upper bound of any $x_i$ is below the maximum lower bound, there is no point in considering it. 

For the $t = \text{ReLU}(x)$ function, \citet{tjeng2017evaluating} introduce the following encoding, which we implement:
\begin{subequations} 
    \begin{align}
    t &\geq 0 \\
    t &\geq x \\ 
    t &\leq uz \\ 
    t &\leq x - l(1-z) \\
    z &\in \{0,1\}
\end{align} 
\end{subequations}
where $[l,u]$ are bounds on $x$ and $z$ is a binary variable.

Finally, we straightforwardly derive a new encoding for absolute value based on the same ideas.
The classic ``big-M'' encoding for absolute value~\citep{aps2018mosek} $t = |x|$ is
\begin{subequations} 
    \begin{align} 
    x &= x^+ - x^-,\\
    t &=x^+ + x^- \\ 
    0 &\leq x^+, x^- \\
    x^+ &\leq Mz \\ 
    x^- &\leq M(1-z) \\ 
    z &\in \{0, 1\}
\end{align} 
\end{subequations}
where $x^+$ represents the positive part of $x$, $x^-$ represents the negative part, and $z$ is a binary variable indicating whether $x > 0$.
Our tighter encoding is:
\begin{subequations} 
    \begin{align}
    x &= x^+ - x^-,\\
    t &=x^+ + x^- \\ 
    0 &\leq x^+ \leq \max(u,0) \\
    0 &\leq x^- \leq \max(-l,0) \\
    x^+ &\leq uz \\
    x^- &\leq |l|(1-z) \\
    z &\in \{0, 1\}
\end{align} 
\end{subequations}
where $[l,u]$ are bounds on $x$.
\newpage
\section{Airplane Fuel Gauge Model Details}
\label{sec:fuel_gauge_model}
The forward measurement model mapping fuel mass to noisy pressure measurements and noisy acceleration measurements was described in brief in the body of the paper, and is described in more detail here.

The state $\vect{x}$ consists of the current mass of fuel in the tank $m$, roll $\phi$ and pitch $\theta$ of the aircraft, and ambient air pressure $P_a$. The forward measurement model computes readings at 9 pressure sensors in the fuel tank as well as the three-dimensional acceleration of the aircraft, which all have associated noise.
The problem is stated as follows:
\begin{align}
    \vect{x} &= [m,~\phi,~\theta,~P_a] \\
    \vects{\nu} &= [\nu_{p_1}, ..., \nu_{p_9}, \nu_{a_x}, \nu_{a_y}, \nu_{a_z}] \\
    \vect{y} &= [p_1, ..., p_9, a_x, a_y, a_z] \\
    \vect{z} &= [m] 
\end{align}
The inverse model attempts to recover the fuel mass $m$, and we want to quantify the error between the neural network estimate of the fuel mass $\hat m$ and its true value: 
\begin{equation}
\max_{\vect{x}}|m-\hat{m}|
\end{equation}
where $|\cdot|$ denotes absolute value.
The pressure reading at a given sensor is determined by the height of fuel above the sensor:
\begin{align}
&p = \max(h_f - h_s, 0) \cdot \frac{\rho a}{c} + P_a + \nu_p
\end{align}
where $h_f$ is the fuel height, $h_s$ is the sensor height, $\rho$ is the fuel density, $a$ is the acceleration magnitude, $c$ is a scaling factor, $P_a$ is ambient air pressure, and $\nu_p$ is Gaussian noise distributed as $\nu_p \sim \mathcal{N}(0, \sigma_p)$. 
The fuel height at a given point in the tank and the three-dimensional position of a given pressure sensor are both functions of the roll $\phi$ and pitch $\theta$ of the aircraft. The fuel height is also a function of the volume $v$ of fuel in the tank:
\begin{align}
    h_f = (1-0.5) f_{lo}(v, \theta, \phi) + 0.5 f_{hi}(v, \theta, \phi) \\
    h_s = (1-0.5) g(\theta, \phi, \vect{p_l} ...) + 0.5 g(\theta, \phi, \vect{p_h} ...)
\end{align}
where $f_{lo}$ is a mapping from $(v, \theta, \phi)$ to fuel height when the wings are unloaded and analogously $f_{hi}$ is a mapping from $(v, \theta, \phi)$ to fuel height when the wings are fully loaded.
Similarly, $\vect{p}_l = (x_l, y_l, z_l)$ is the position of the sensor when the aircraft is at nominal attitude ($\theta=0$, $\phi=0$) with wings unloaded, and $\vect{p}_h = (x_h, y_h, z_h)$ is the position of the sensor at nominal attitude with wings loaded. 
The function $g$ maps the pressure sensor coordinates $\vect{p}_l$ and $\vect{p}_h$ from nominal attitude to an arbitrary roll $\phi$ and pitch $\theta$:
\begin{equation} \label{eq:g}
    g(\theta, \phi, x, y, z) = x \sin(\phi) - y \sin(\theta)\cos(\phi) + z \cos(\theta) \cos(\phi)
\end{equation}

The acceleration at a given attitude is computed:
\begin{align}
    a_x &= a\sin(\phi) + \nu_{a_x}\\ 
    a_y &= -a\sin(\theta)\cos(\phi) + \nu_{a_y}\\
    a_z &= a\sqrt{(1 - \sin(\phi)^2 - (\sin(\theta) \cos(\phi))^2)} +\nu_{a_z} \label{eq:accel}
\end{align}
where $a$ is the magnitude of total acceleration and $(\nu_{a_x}, \nu_{a_y}, \nu_{a_z})$ are independent Gaussian noise variables each distributed $\nu_a \sim \mathcal{N}(0, \sigma_a)$. 
The magnitude of total acceleration is fixed at $a=\SI{10}{\meter\per\second\squared}$.

\newpage
\section{Fuel Gauge Experimental Details}\label{sec:comp_deets}
Fuel gauge verification experiments were run across 5 servers with a collective 148 physical CPU cores and 296 logical CPU cores. Fourteen Julia threads were used, but each call to the optimizer calls the BLAS library which was allowed to spawn a number of threads up to the number of physical CPU cores on each server. The processors of the five machines used were all Intel(R) Xeon(R) CPUs of different models. See details below.
\begin{table}[h!]
\centering
\begin{tabular}{@{}lc|c|c@{}}
\toprule
\textbf{Xeon Processor}          & \textbf{Physical Cores} & \textbf{Logical Cores} & \textbf{Julia Threads} \\
\midrule
 E5-2650 v3 @ 2.30GHz & 20 & 40                       & 3             \\
 E5-2699 v4 @ 2.20GHz & 44 & 88                      & 4             \\
 E5-2690 v4 @ 2.60GHz & 28 & 56                      & 3             \\
 E5-2690 v4 @ 2.60GHz & 28 & 56                      & 1             \\
 E5-2690 v4 @ 2.60GHz & 28 & 56                      & 3 \\
\bottomrule
\end{tabular}
\end{table}
\newpage
\section{Additional Fuel Gauge Results}
\begin{figure}[h!]
            \centering
            \input{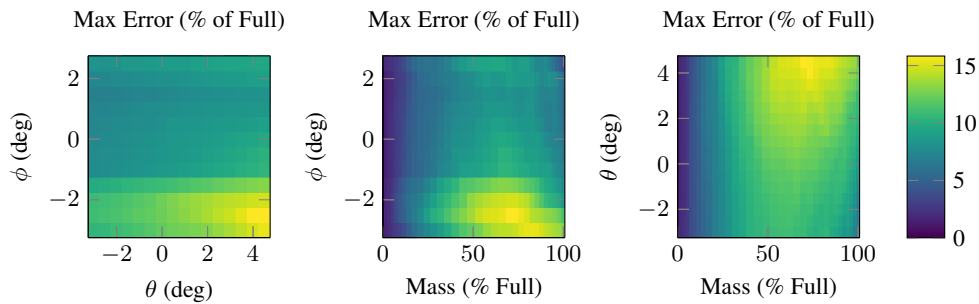}
            \caption{Estimation error with $3\sigma$ noise across different slices of the state space.}
            \label{fig:all_heatmaps}
        \end{figure}

\cref{fig:all_heatmaps} shows the fuel mass estimation error plotted with respect to different system variables. 
For the left plot a maximum was taken over mass values, for the middle plot a maximum was taken over pitch ($\theta$) values and for the right plot a maximum was taken over roll ($\phi$) values.
        
\begin{figure*}[h!]
            \centering
            \input{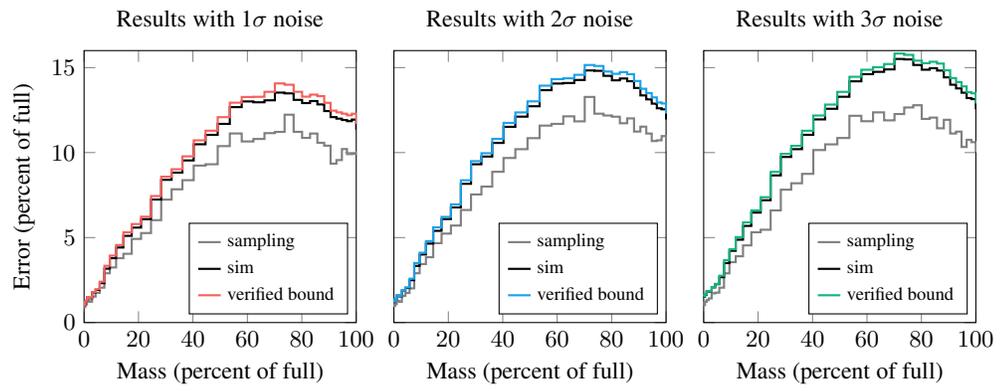}
            \caption{Verified upper bounds are compared with error estimates obtained from sampling. There is a tradeoff between the desired confidence level in the verified upper bound and the tightness of the bound. A more confident upper bound will be looser.}
            \label{fig:nn_sampling_three}
\end{figure*}

\Cref{fig:nn_sampling_three} displays verified upper bounds for three noise domains of the 12 random variables in the input: 1$\sigma$, 2$\sigma$, and 3$\sigma$.
This gives upper bounds corresponding to $1.02 \%$, $57.12 \%$, and $96.81 \%$ of the probability mass of the error distribution, respectively.
The risk threshold is ultimately a user-chosen parameter and a verified upper bound considering higher noise levels could also be computed.
Increasing the noise level increases the conservatism of the verified upper bound and the time required to compute it.
The time taken to compute verified upper bounds on the error for each level of noise on the inputs is shown in \cref{tab:time}. Experiments where run on 
148 physical cores (see \cref{sec:comp_deets}).

The time taken for each noise level is shown in \cref{tab:time}.

        \begin{table}[h]
        \centering
        \begin{tabular}{@{}lrrr@{}} 
        \toprule
                & \multicolumn{1}{p{1cm}}{\centering \textbf{\scriptsize Total} \\ \textbf{\scriptsize Time (hr)} } 
                & \multicolumn{1}{p{1cm}}{\centering \textbf{\scriptsize Time per} \\ \textbf{\scriptsize Cell (s)} } 
                & \multicolumn{1}{p{1cm}}{\centering \textbf{\scriptsize Probability} \\ \textbf{\scriptsize Mass}  } \\
        \midrule
        \scriptsize $0\sigma$ & \scriptsize 0.718 & \scriptsize 0.345  & \scriptsize - \\
        \scriptsize $1\sigma$ & \scriptsize 1.172 & \scriptsize 0.563  & \scriptsize 0.0102 \\
        \scriptsize $2\sigma$ & \scriptsize 1.344 & \scriptsize 0.646  & \scriptsize 0.5712 \\
        \scriptsize $3\sigma$ & \scriptsize 1.520 & \scriptsize 0.731  & \scriptsize 0.9681 \\
        \bottomrule
        \end{tabular}
        \caption{Time taken to verify the fuel gauge case study. }
        \label{tab:time}
        \end{table}

\end{document}